# Developing ANFIS-PSO Model to Predict Mercury Emissions in Combustion Flue Gases

Shahaboddin Shamshirband[1,2], Masoud Hadipoor[3], Alireza Baghban[4], Amir Mosavi[5,6,*], Jozsef Bukor [7], Annamaria R. Varkonyi-Koczy [5,7]

[1] Department for Management of Science and Technology Development, Ton Duc Thang University, Ho Chi Minh City, Vietnam;

[2] Faculty of Information Technology, Ton Duc Thang University, Ho Chi Minh City, Vietnam

[3] Department of Petroleum Engineering, Ahwaz Faculty of Petroleum Engineering, Petroleum University of Technology (PUT), Ahwaz;

[4] Department of Chemical Engineering, Amirkabir University of Technology (Tehran Polytechnic), Mahshahr Campus, Mahshahr;

[5] Institute of Automation, Kando Kalman Faculty of Electrical Engineering, Obuda University, Budapest 1034, Hungary

[6] School of Built the Environment, Oxford Brookes University, Oxford OX30BP, UK;

[7] Department of Mathematics and Informatics, J. Selye University, Komarno 94501, Slovakia;

**Abstract:** Accurate prediction of mercury content emitted from fossil-fueled power stations is of utmost importance for environmental pollution assessment and hazard mitigation. In this paper, mercury content in the output gas of power stations' boilers was predicted using an adaptive neuro-fuzzy inference system (ANFIS) method integrated with particle swarm optimization (PSO). The input parameters of the model include coal characteristics and the operational parameters of the boilers. The dataset has been collected from 82 power plants and employed to educate and examine the proposed model. To evaluate the performance of the proposed hybrid model of ANFIS-PSO model, the statistical meter of MARE% was implemented, which resulted in 0.003266 and 0.013272 for training and testing, respectively. Furthermore, relative errors between acquired data and predicted values were between -0.25% and 0.1%, which confirm the accuracy of the model to deal nonlinearity and representing the dependency of flue gas mercury content into the specifications of coal and the boiler type.

## 1. Introduction

The huge dependency on fossil fuels in the production of energy to support industries, mobility, and urbanization have dramatically increased air pollution worldwide [1-3]. The population growth, industrialization, climate change, and the ever-growing urbanization further are accelerating the severe effect on the air quality and emissions [4-6]. The air pollution is known as a profound contributor to human mortality and potential danger to the environment and ecological systems. [2,7,8]. Thus, intelligent monitoring of the air pollutants is of utmost importance to maintain acceptable air quality for well-being [9-12].

Among the numerous industrial pollutants, mercury contamination has been identified as one of the most acute air pollutants produced by conventional fossil-fueled power stations [13-16]. Mercury contamination can cause significant ecological hazard with a considerable effect on human well-being around the world [17-20]. As a lethal and hugely volatile metal, mercury can cause contamination of the surface streams and lakes, as well as groundwater [21]. It is the most dangerous hazard for infants and young adults as it influences the central nervous system, causing severe illnesses [22]. Previous studies, e.g. [7-11], report that a substantial amount of mercury outflows to the earth comes from coal-fired power plants. In 2010, roughly 1960 tone/year mercury outflowed to the air from various sections worldwide [23]. Coal-burning had a share of 24%, which is a relatively high share [24]. Power plants are in charge of around 33% Mercury outflows, and this type of emission is caused by human beings [25], and Elemental mercury emission is about 20-50% of mercury emissions which originate from combustion of coal [26,27]. Nowadays, mercury emission from coal consumption has become a global concern [12,13,14]. In 2006, total coal consumption in China was about 40.1% of world consumption, which is equivalent to 1238.3 million tons of oil [28]. Thus, some studies suggest that the amount of mercury emission is more likely to increase during the next years because of more uses in developing countries [29]. The environmental protection agency of the United States of America announced mercury as one of the most dangerous air pollutants. In 1999, an approximated amount of 45 tons of mercury outflows from coal-consuming plants to the environment (Alto 2000). The developing worry of this contamination in the U.S has incited government and specialists to start endeavors to recognize, estimate, and cut off on the anthropogenic emissions. As a result of the absence of cost-effective, promptly accessible and efficient practical control methodologies in the U.S, discharge of this dangerous contaminant from coal-consuming boilers are not basically under control. It gets worse when the greater part of power supply in a big country such as the United States originates from utility boilers that use coal (EPA 2001) and furthermore about 70% of electricity power in China is produced by burning coal, in which 50% of this coal is burned in coal-based power plants [30-32].

In 1998, paying attention to the enormous potential for environmental dangers, EPA proposed a request to ask coal-consuming plants to publish information on the amounts of mercury contaminant outflows from their systems. This request was designed to gather information in three primary stages precisely. The first and principal stage was intended to collect all standard data on coal-burning power plants around the U.S. Afterward, as the second stage of the program, analyzed feed data at the entrance of every plant during a year were collected. Eventually, in the third phase, EPA picks 84 out of 1084 plants to gather data of mercury emission in some specified points within the selected plants. This selection was based on some statistic activities on the feed specifications and also the operational structure of each plant. Obtained data from the third phase of the program was evaluated. Representing correlations were developed to predict the emission of mercury in each plant concerning coal qualities and operating conditions. It was found that the best input data were characteristics of coal, for example, the concentration of mercury, heating value, chlorine sulfur, operating parameters such as temperatures and pressures and also yield parameters in boilers such as the amount of mercury oxidation.

Recently, intelligent and data-driven methods have become increasingly famous for the prediction of air pollution [33-37]. Among them, machine learning methods have been reported to deliver higher performance, in term of accuracy, robustness, and lower computation power in dealing with uncertainties and big data [38-41]. Several surveys report that ensemble and hybrid models are the future trends in machine learning due to their optimized algorithms for higher efficiency [42-48]. Hybrid machine learning models are shown to deliver higher perfoamnce in air pollution modeling and prediction [49-54]. However, the application of hybrid machine learning in modeling mercury emission has been limited and proposes a research gap. Consequently, this paper aims to propose one of the newly advanced hybrid models to predict mercury emission. This paper deal with the problem of the prediction of mercury emission in nature using an outstanding and new method of ANFIS-PSO model. Authors have developed a really powerful model empowered by PSO optimization algorithm. Data were obtained from the literature survey and then analyzed using the proposed model.

The rest of the paper is organized as follows. In the second section a review of the literature is provided and the reason behind selecting the ANFIS-PSO. In the third section, model development is described. The results and the conclusions are presented in the sections fourth and fifth, respectively.

## 2. Previous Investigations

Artificial intelligence approaches are powerful tools to model and predict the parameters of air pollution, including mercury emissions through finding correlations between variables [37,55-58]. Among the Artificial intelligence approaches, the machine learning methods are particularly known as the powerful algorithms in delivering insight into the nonlinear relationship between parameters [56,59,60]. A deep understanding of the power plant is needed to control the amounts of mercury emissions [61-63]. Therefore, an accurate estimation of emission is of utmost importance to control and reduce mercury emission [64]. Numerous investigations were published in the literature regarding applications of artificial intelligence approaches [65-69]. Computational intelligence has been both used to predict the amount of mercury emission and also to model the elimination of elemental mercury from boilers' outlet gas [56]. Dragomir and Oprea [70] present a multi-agent prediction tool for intelligent monitoring of the pollutants on the power plants. They used a model based on neural networks to predict the amount of $SO_2$, $NO_x$, particulate matters (PMs), and mercury emissions. Jensen et al. [71] presented a study on the relationship between mercury in the flue gas and coal specifications and the type of boiler using a multilayer perceptron model. They derived an accurate model with a correlation coefficient of 0.9750. Antanasijevic et al. [72] developed a prediction model using neural networks and genetic algorithm (GA) to accurately calculate the amount of PM10 emissions for up to two years ahead. Zhao et al. [73] used support vector machine to develop a model which provided better performance and accuracy. In 2016, Wang et al. [74] worked on the application of GA-back propagation (GA-BP) for predicting the amount of mercury component in flue gases of 20 different coal-fired boilers. Correlation coefficient training data points was as high as 0.895, and they showed that GA-BP is a promising method for this goal. Li et al. [75] employed computational intelligence approach to cut off on the elemental mercury in coal-fired boilers, and finally, they found that the increment of capture efficiency can approximately improve up to 15%.

Although the application of machine learning for prediction of pollutants and mercury emissions is well established within the scientific communities, the potential of the novel machine learning models (e.g., ensembles and hybrids) is still not explored for mercury prediction. In particular, a wide range of novel hybrid machine learning methods has been recently developed to deliver higher accuracy and performance [47,76,77]. For instance, the hybrid model of ANFIS-PSO which is an integration of adaptive neuro-fuzzy inference system (ANFIS) and particle swarm optimization (PSO) has shown to deliver promising results [78]. The hybrid model of ANFIS-PSO (also known as PSO-ANFIS) has been appeared in the works of Catalao et al. [79,80], in early 2011 for prediction of wind energy and electricity pricing prediction. Since then this method has been used in various applications, e.g., load shedding, electricity prices forecasting, hydrofoil, travel time estimation, prediction of viscosity of mixed oils, matrix membranes modeling, wax deposition, electric power forecasting, asphaltene precipitation, prediction of density of bitumen diluted with solvents, heating value of biomass, predict interfacial tension of hydrocarbons and brine, prediction of gas density, forecasting oil flocculated asphaltene, biodiesel efficiency, Biomass higher heating modeling, prediction of property damage, and solar radiation forecasting [81-97]. The generalization ability, higher accuracy, speed, and ease of use have been reported as the main characteristics of ANFIS-PSO. Therefore, this hybrid method has been identified as a suitable candidate for modeling mercury emission. Concequnetly, the aim of the present study is set to find a reliable relationship between elemental mercury in the output gas, the specification of feed, and the type of boilers by utilizing an ANFIS-PSO based approach.

**3. Model development**

The description of the hybrid model of ANFIS-PSO is presented in [37]. Note that, when there is not enough data on the detailed information of an operating power plant, it is extremely difficult to build a precise model to predict the amount of mercury outflow. In the present study, an endeavor has made to develop a model to predict mercury outflows from boilers at some specified testing locations. In this kind of locations, every single factor that may influence the mercury discharge is considered and incorporated into the model. A total number of 82 data points were gathered from literature to train and evaluate the model [71]. The concentration of mercury in the inlet feed, ash content, chlorine content, the heating value of coal, sulfur content, and temperature were chosen as the most effective variables. This data bank comprises a total number of 82 data points, from which 75% were used as training, and the rest of them were exploited testing samples. In the developed ANFIS model, six previously mentioned parameters were considered as input parameters, and the elemental mercury emission was selected as the target variable. Furthermore, the PSO algorithm was used to find the optimized Gaussian membership function parameters of the proposed ANFIS model.

The method of ANFIS is proposed by Jang [98,99] and is a versatile and very intelligent hybrid system. ANFIS approach can be expressed as a complete collaboration between computing activities and neuro-fuzzy system [100]. This method integrates natural and neural networks and uses their strength into its advantage. Such methodology exploits back-propagation calculation from the information gathering process to make the essential basics of the fuzzy system.

Learning capability of the proposed network structure is a result of the combination of different types of neural network's capabilities. ANN and fuzzy systems are combined to form a firmly structured network, neuro-fuzzy systems. In addition to that, they allow a really simple transformation of the whole system into if-then rules which is one of the greatest advantage of these networks [6]. Its framework is related to an arrangement of fuzzy IF-THEN rules which have learning ability to estimate nonlinear functions. Neuro-fuzzy inference systems have been used in many research activities and there is no doubt that neural-fuzzy combined systems are powerful in many fields. As it can be concluded from their name, adaptive networks are constructed from nodes and directed paths, and all I/O values can be modifiable by different sets of parameters which are defined in the architecture of these networks [6]. On the other hand, ANFIS systems can utilize a various range of algorithms to decrease the final errors of the training phase. For instance, the gradient descent approach could be combined with the least square method to optimize the effectiveness of the searching process for the best parameters. The convergence rate of hybrid approaches is too faster because they lower the dimensions of searching space in the backpropagation process [101].

Basics of the ANFIS method are approximately similar to a fuzzy system developed by Takagi-Sugeno-Kang [102,103]. In reverse spread learning capability of the ANFIS method, which is based on the calculation of derivatives of squared errors in a backward manner from output nodes to the input ones, this method constructs and utilizes robust learning methodology based on gradient least-squares approach. To determine the consequence factors in the forward section, the least square approach is utilized. Then the preset parameters will reset by gradient descent in the regressive advance [104]. The adaptive network is constructed of five layers. Figure 1 shows these layers, their nodes, and connections with the assumption of two inputs to the fuzzy inference system expressed by "x" and "y" and a single output of "f". As an explanation about the configuration of ANFIS, it must be noted that two fuzzy 'if-then' rules are utilized which they follow Sugeno FIS as:

$$f_1 = P_1 + q_1 y + r_1, \qquad \text{assume } x=A_1, y=B_1$$
$$f_2 = P_2 + q_2 y + r_2, \qquad \text{assume } x=A_1, y=B_1$$

Fuzzification layer, which is the first layer of the structure produces all membership grades for each variable. Node functions in this layer can be defined as follows:

$$O_{1,i} = \mu_{Ai}(x), \qquad i = 1,2 \qquad (1)$$

$$O_{1,j} = \mu_{Bj}(x), \qquad j = 1,2 \qquad (2)$$

Memberships of a fuzzy set are ($A_i$, $B_i$) and $O_{1,i}$ represents the resulted value from the $i^{th}$ node of the first layer. The input signals are generated by the nodes of layer 2.

$$O_{2,i} = W_i = \mu_{Ai}(x) \times \mu_{Bi}(x), \qquad i = 1,2 \qquad (3)$$

The nodes of the third layer are used to compute the following parameter:

$$O_{3,i} = \overline{w} = \frac{W_i}{W_1 + W_2}, \qquad i = 1,2 \qquad (4)$$

Where $W_i$ is ruled firing strengths of node i which has a normalized firing strength of $\omega_i$. Results of layer four can be written as follows:

$$O_{4,i} = \overline{w} f_i = \overline{w}_i (P_i + q_i y + r_i), \qquad i = 1,2 \qquad (5)$$

In this notation $p_i$, $q_i$, and $r_i$ are called consequent parameters. Eventually, the general output can be defined as follows, which is calculated in the nodes of layer 5:

$$O_{5,i} = \sum_{i=1}^{2} \overline{w}_i f_i = \frac{W_1 f_1 + W_2 f_2}{W_1 + W_2}, \tag{6}$$

So the final output of the ANFIS can be written as follows:

$$Z = \frac{W_1}{W_1 + W_2} f_1 + \frac{W_2}{W_1 + W_2} f_2 + \ldots + \frac{W_n}{W_{n-1} + W_n} f_n, \tag{7}$$

In these networks, the combination of back propagation and the least square approaches will result in faster convergence and more precise values and as a consequence, a better learning ability. Least square is very useful in determining the optimized values of the fourth layer which are called consequent parameters. In addition to that, the premise parameters which are located in the first layer must be optimized in order to define the best shape of membership functions [18]. These parameters will be optimized with respect to the output errors which must be minimized using the back propagation method [105].

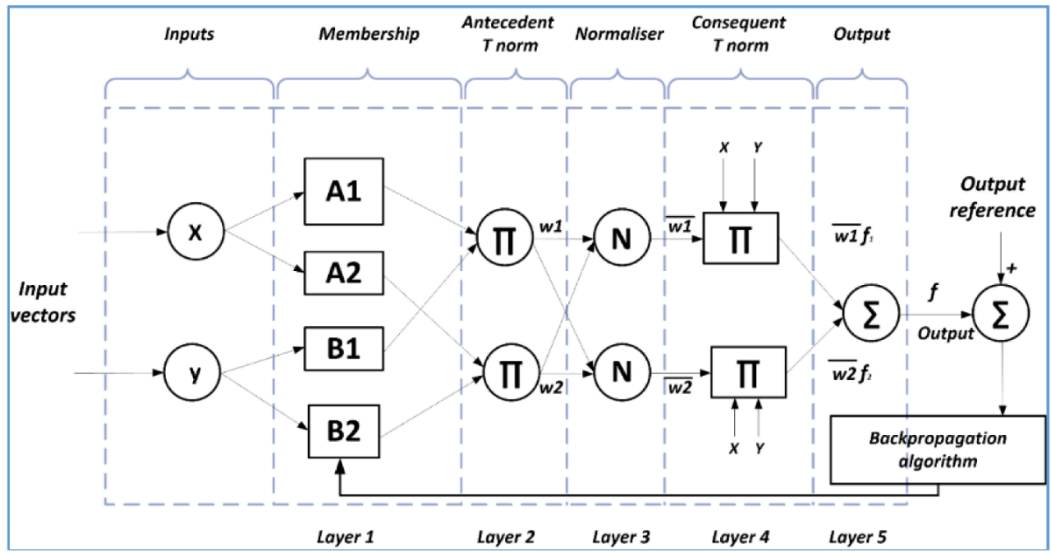

**Figure 1.** A schematic view of the ANFIS intelligent system.

ANFIS has shown promising results in a wide range of applications for developing prediction models [106-108]. However, optimization of the model parameters can dramatically improve the quality and accuracy of modeling [78]. For that matter, a huge number of optimization methodologies, such as PSO, are available to reinforce the parameters and answers of the ANFIS system [109]. PSO is extraordinary compared to other approaches with the end goal of optimization. This study takes the benefits of this algorithm.

Particle swarm optimization method has been inspired by birds behavior seeking food [110,111]. In this model, particles update their places and pathways based on their and others information; so it was proposed that the particle possess a memory function. The optimization process is based on competition and collaboration between particles. When PSO is used to solve optimization problems, one can follow the particles state by their pathways, and velocities.

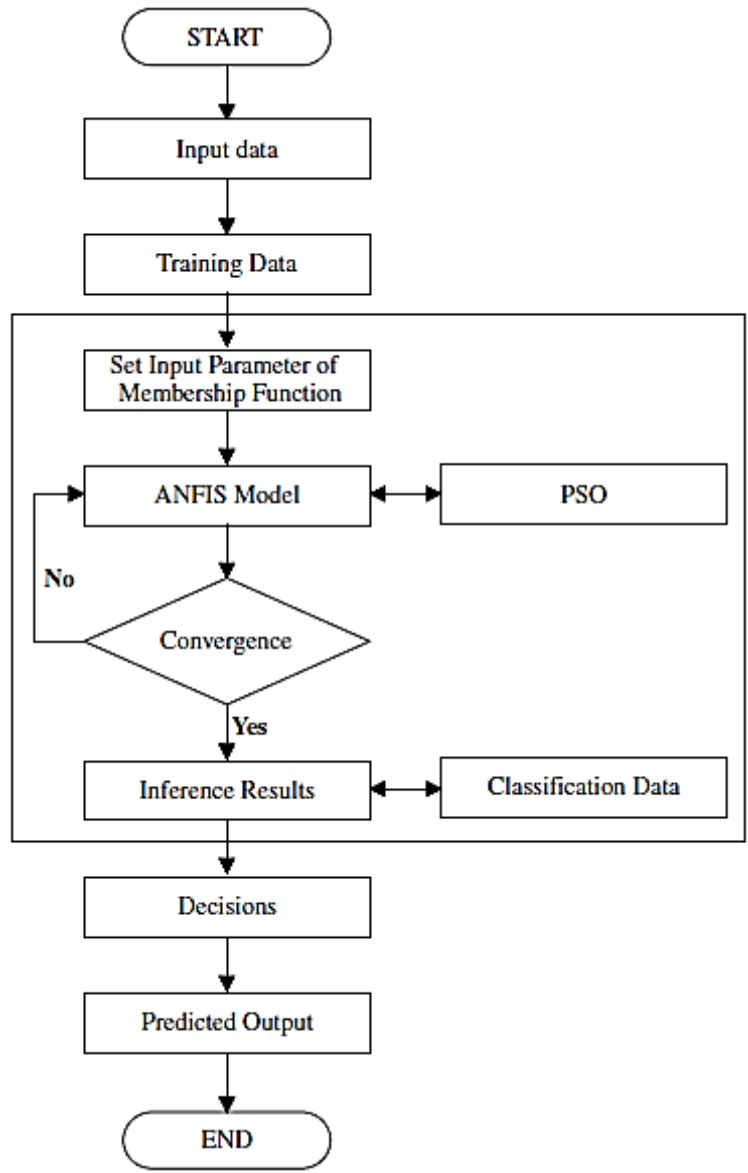

**Figure 2.** Diagram of optimized ANFIS structure by PSO algorithm (ANFIS-PSO)

Three vectors $X_i$, $V_i$, $Pbest_i$ are introduced to explain the properties of a particle: $X_i$ is the current place; $V_i$ the current speed; $Pbest_i$ the best spatial placement sought by the particle and $gbest_i$ is the optimal solution searched by the whole group of particles.

The position and pathway of the particle will be updated gradually, based on the following formula:

$$v(k+1)=v(k)+c_1 \text{rand}(0,1)\times[\text{pbest}(k)-\text{persent}(k)]+c_2 \text{rand}(0,1)\times[\text{gbest}(k)-\text{persent}(k)], \quad (8)$$

$$\text{present}(k+1) = \text{present}(k) + v(k+1), \quad (9)$$

Where, v( ) is particle speed in *k*th and *k+1*th iterations; present ( ) is particle position; *c*1, *c*2 are learning constants which are greater than zero, and a random number between [0,1] is denoted using rand( ). Formula (7) represents the updating process of the particle's speed, which includes a particle's historical velocities and personal and global best positions [112].

Diagram of ANFIS-PSO approach was shown in Figure 2. The further detailed description and information on ANFIS-PSO have been given by Basser et al. [78]. Accordingly, the developed model for an estimation case is created based on the following three steps:

- Dataset is partitioned into different clusters by kernel-based clustering approaches

- The cluster centers obtained from clustering are applied to create the fuzzy rule base of ANFIS

- The resulting ANFIS model is trained through the PSO method.

## 4. Results

The amount of mercury emission was estimated using an ANFIS approach. Emission of mercury into the environment generally is a strong function of mercury six previously mentioned variables. We used MATLAB software to construct our model. A Gaussian function was used to optimize the parameters. In addition to that, the total number of 10 clusters were utilized in the ANFIS hybrid system. Optimization was conducted on a total number of parameters that were determined by:

$$N_T = N_c N_v N_{mf}, \qquad (10)$$

Where the number of parameters for undergoing optimization is denoted by $N_T$, and $N_{mf}$ is used to show the number of Gaussian membership functions that are used, $N_v$ and $N_c$ show how many variables, and clusters are used in the model, respectively. It is noteworthy to state that in this study, two membership functions, seven input and output variables, and 10 clusters are used. Eventually, using a PSO algorithm, optimization was conducted for 140 tuning parameters. As is shown in Figure 3, to evaluate the functionality of the PSO algorithm, a root means square error (RMSE) analysis was used. Results show that in a total number of 1000 iterations, the minimum value of RMSE is touched. Figure 4 indicates train membership function parameters for each input variables. It is seen that the results of the presented model are in good agreement with the obtained data, which is the result of great learning capability of the developed ANFIS model. Figure 5 illustrates the obtained data of mercury emissions versus the test and training of ANFIS hybrid system.

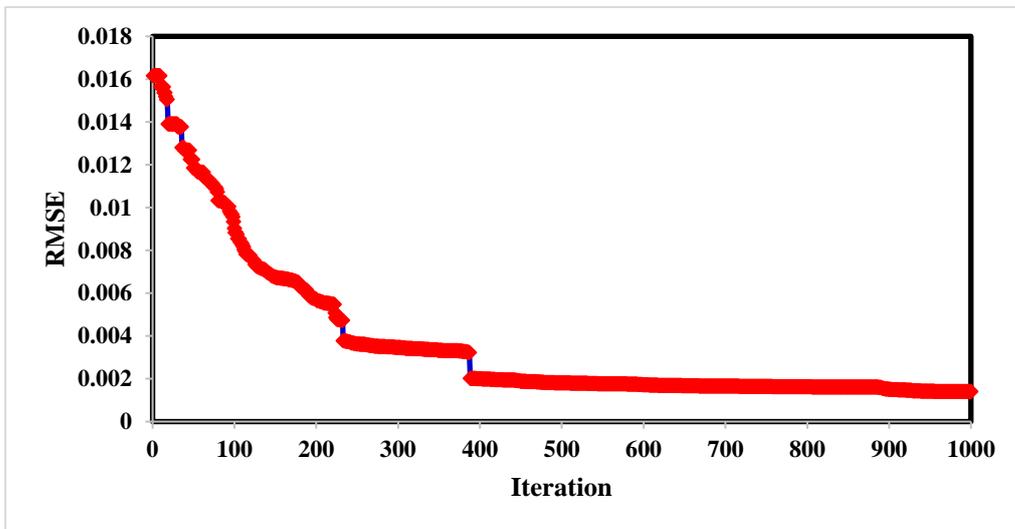

**Figure 3.** Root mean square errors versus the number of iterations.

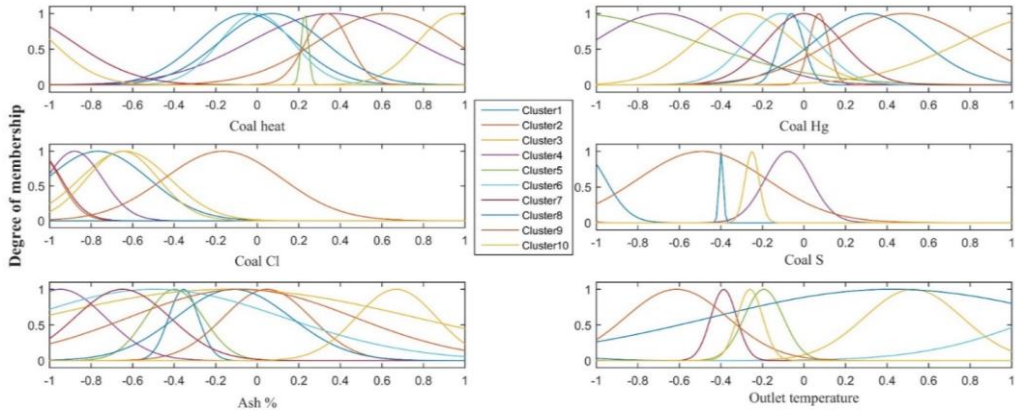

**Figure 4.** Trained membership function parameters

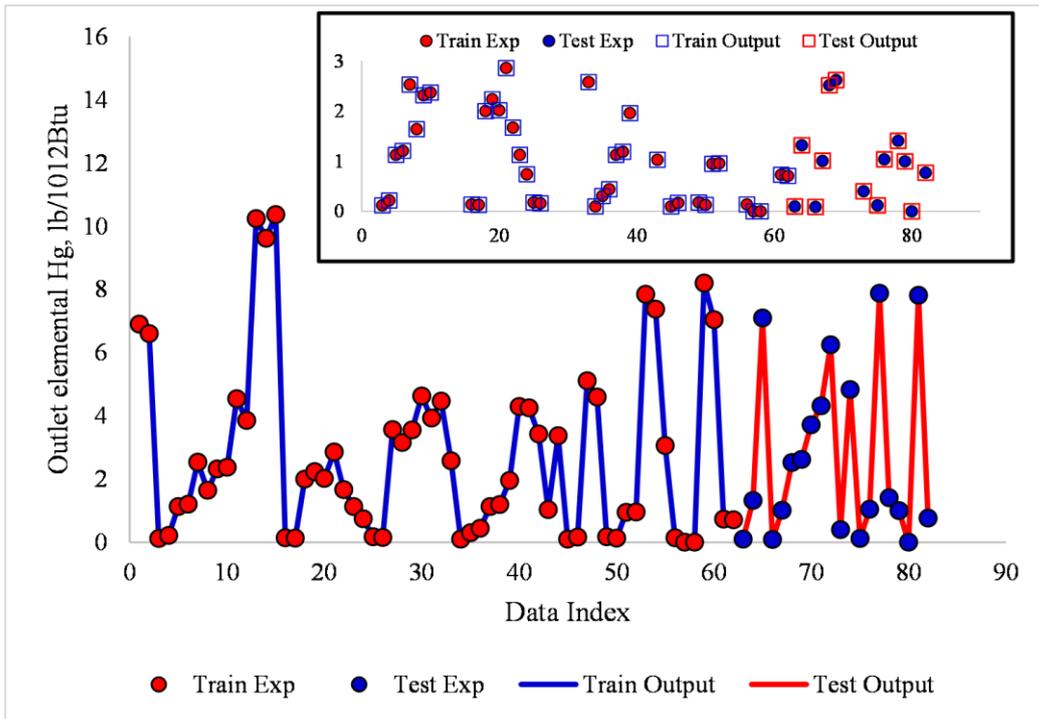

**Figure 5.** Obtained data form plants and ANFIS values for mercury emissions in the stages of training and testing.

A scatter diagram is a straightforward statistical technique used to indicate a relationship between two parameters. It is frequently joint with a simple linear regression line used to fit a model between the two parameters. As is shown in Figure 6, actual and predicted mercury emissions are located on a straight line with an approximate slope of 1 (45º line) which indicates that the obtained information and ANFIS predicted ones are in good agreement. The obtained cross-fit line in both test and training data sets have an $R^2$ Equal to 1, which shows the accurateness of the model. To compare the results of the model and evaluate its precision, the method of mean absolute relative error is used. For training and testing steps, using mean absolute relative error percentage (MARE %) method, percentage values of 0.003266 and 0.013272 are calculated, respectively. Resulted relative deviations are presented in Figure 7. Low relative deviations are observed due to accurately-predicted values. Different statistical analyses were also presented in Table 1 for the suggested model.

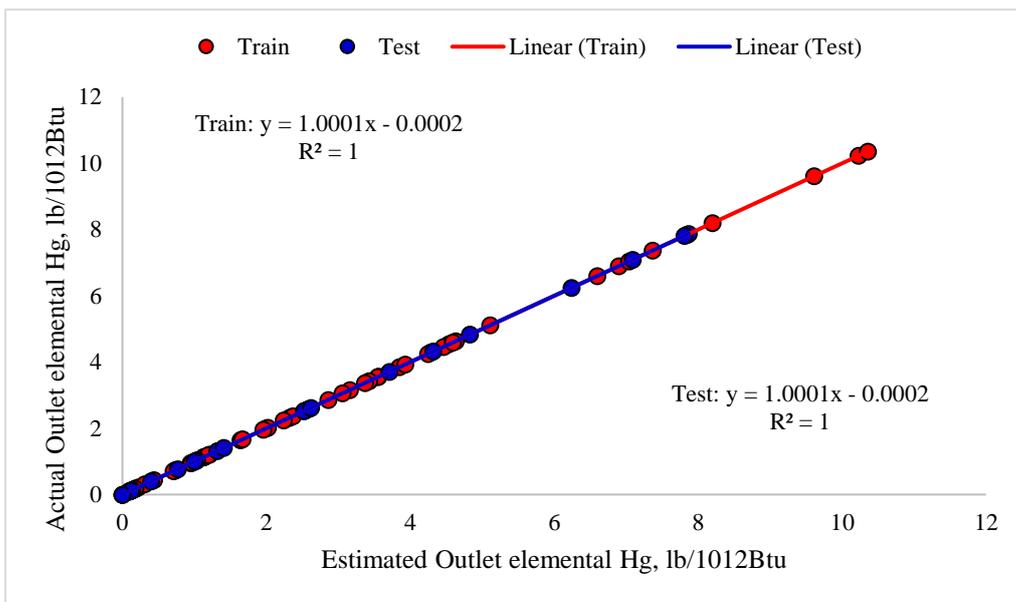

**Figure 6.** Regressions derived between estimated and collected data of mercury emissions.

Table 1: Statistical analysis of the model for all phases

|  | Train | Test |
|---|---|---|
| $R^2$ | 1.000 | 1.000 |
| MSE | 1.40E-07 | 1.39E-07 |
| MRE (%) | 0.037 | 0.044 |

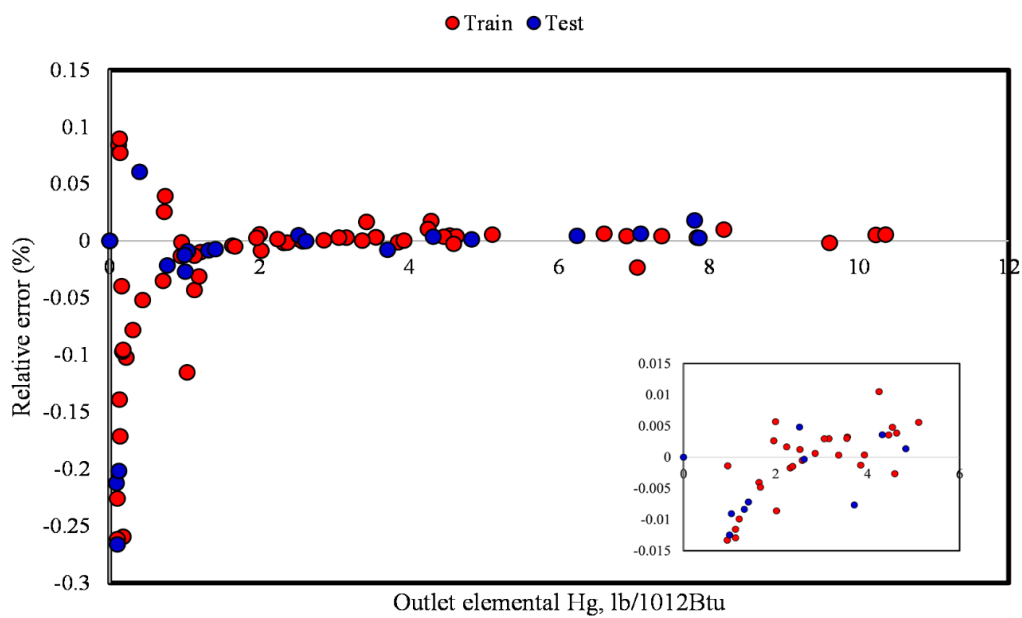

**Figure 7.** The deviation between the obtained data from plants and predicted mercury emissions.

## 5. Conclusions

Emission of mercury is known as one of the most perilous environmental contamination. In this study, a comprehensive literature review was done, and a predictive model was built to predict the amount of mercury emission based on the characteristics of the coal supply, operational conditions, and so forth. The presented model is based on the ANFIS system, which utilizes a PSO algorithm to estimate the amount of mercury emission to the environment. Data from 82 power plants have been used to train and develop the ANFIS model. Optimized corresponding membership functions for each of the clusters have presented separately. Between iterations 0 and 230, a dramatic and very fast decrease in the figures of RMSE values has been seen and the figures stayed relatively stable afterward which reflects that the speed of convergence is relatively high. Percentages of MARE for training and testing were 0.003266 and 0.013272, respectively. Additionally, figures of MSE for the training and testing sections were 1.4E-07 and 1.39E-07 and the resulted values for MRE% for training and testing sections of the modeling were 0.037 and 0.044, respectively. Furthermore, relative errors between acquired data and predicted values were between -0.25% and 0.1%, which confirm the accuracy of PSO-ANFIS model. It was seen that for both training and testing parts, the coefficient of determination was calculated to equal to unity, which reflects the accuracy of the proposed ANFIS-PSO based model.

**Author Contributions:** conceptualization, modeling, data curation, data analysis and analyzing the results, S.S., M.H., A.B., and A.M.; machine learning and soft computing expertise, S.S., M.H., A.B., A.M., J.B., and A.R.V.K.; mathematics expertise, M.H., A.B., and J.B.; management, database, writing, administration and methodology, M.H., A.B.; visualization, M.H., A.B.; supervision, resources, software, expertise, revision, funding and verifying the results, J.B., and A.R.V.K.

**Conflicts of Interest:** The authors declare no conflict of interest.

**Acknowledgment:** This publication has been supported by the Project: "Support of research and development activities of the J. Selye University in the field of Digital Slovakia and creative industry" of the Research & Innovation Operational Programme (ITMS code: NFP313010T504) co-funded by the European Regional Development Fund.